# LF2L: Loss Fusion Horizontal Federated Learning Across Heterogeneous Feature Spaces Using External Datasets Effectively: A Case Study in Second Primary Cancer Prediction

Chia-Fu Lin[1] and Yi-Ju Tseng[1,*]

## ABSTRACT

Second primary cancer (SPC), a new cancer in patients different from previously diagnosed, is a growing concern due to improved cancer survival rates. Early prediction of SPC is essential to enable timely clinical interventions. This study focuses on lung cancer survivors treated in Taiwanese hospitals, where the limited size and geographic scope of local datasets restrict the effectiveness and generalizability of traditional machine learning approaches. To address this, we incorporate external data from the publicly available US-based Surveillance, Epidemiology, and End Results (SEER) program, significantly increasing data diversity and scale. However, the integration of multi-source datasets presents challenges such as feature inconsistency and privacy constraints. Rather than naively merging data, we proposed a **loss fusion horizontal federated learning** (**LF2L**) framework that can enable effective cross-institutional collaboration while preserving institutional privacy by avoiding data sharing. Using both common and unique features and balancing their contributions through a shared loss mechanism, our method demonstrates substantial improvements in the prediction performance of SPC. Experiment results show statistically significant improvements in AUROC and AUPRC when compared to localized, horizontal federated, and centralized learning baselines. This highlights the importance of not only acquiring external data but also leveraging it effectively to enhance model performance in real-world clinical model development.

[1] Department of Computer Science, National Yang Ming Chiao Tung University, Hsinchu, Taiwan
[*] Corresponding author: Yi-Ju Tseng (yjtseng@nycu.edu.tw)
This work was funded by National Science and Technology Council, Taiwan, under Grant No. NSTC 114-2221-E-A49-061 and 113-2813-C-A49-009-E

# 1. Introduction

Second primary cancer (SPC) refers to the development of a new primary cancer in a patient who has previously been diagnosed with a different primary cancer (National Cancer Institute [NCI], 2011b). Unlike recurrent cancer or metastasis (NCI, 2011a, 2025), where the original cancer spreads to other parts of the body, SPC arises independently and poses a growing concern for cancer survivors due to the improved survival rate. Early identification of patients at higher risk for SPC is crucial for timely intervention and better clinical outcomes (Sung et al., 2024).

The types of SPCs vary depending on factors such as types of original cancer, lifestyle, treatment modalities (e.g., surgery or chemotherapy), etc. In this study, we focus on predicting SPC among lung cancer survivors, leveraging real-world clinical data. However, our data is collected exclusively from hospitals in Taiwan, which introduces two key limitations. First, the sample size is relatively limited, which may reduce the power of traditional machine learning models (Rajput et al., 2023). Second, because the data originates from a single geographic area, the resulting models may suffer from lack of generalizability when applied to other populations or healthcare systems.

To address this limitation, we incorporate data from additional available sources, specifically the Surveillance, Epidemiology, and End Results (SEER) program (*Surveillance, Epidemiology, and End Results Program*, 2025), which contains a large and diverse cohort of U.S. cancer patients. Integrating SEER data has the potential to significantly improve prediction performance by increasing sample diversity and size. However, directly merging real-world clinical data with datasets from the other sources would introduce challenges due to privacy issues and feature heterogeneity (Soenksen et al., 2022), different sources often collect different sets of attributes, resulting in sparse feature spaces or requiring imputation, which can degrade model performance (Petrini et al., 2022).

Inspired by the previous study (Hong & Tseng, 2022), we explore an alternative route using federated learning (FL) to avoid merging raw data across institutions with the additional benefits of preserving privacy (Zhang et al., 2021). FL allows collaborative model training across multiple data sources without exchanging sensitive patient information. However, traditional horizontal federated learning (HFL) requires that all participating clients share the same feature space (Mammen, 2021), a condition that is not met in our case.

To overcome this limitation, we propose a framework called **Loss Fusion Federated Learning (LF2L)**. This method enables each data source to maintain its unique feature set while contributing to the global model through a shared loss representation. By eliminating the need for feature space alignment, LF2L allows us to fully leverage the predictive value of all available datasets without introducing sparsity or discarding critical information.

## 2. Method

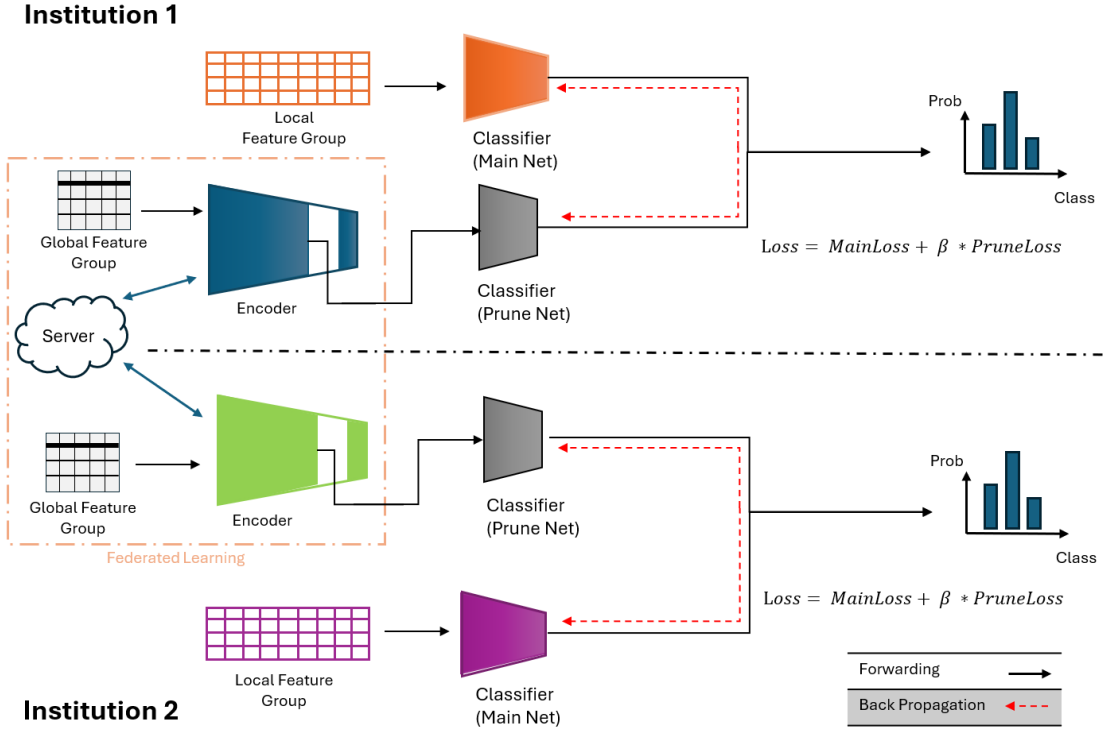

**Figure 1**: The workflow of the proposed Loss Fusion Horizontal Federated Learning Framework (LF2L). β is a learnable parameter, gradually fine-tuning the effectiveness of the federated learning.

**Figure 1** illustrates the proposed LF2L workflow. The process begins by dividing the input features into two groups: global and local feature groups, which may partially overlap (**Figure 2**). The global feature group is used to perform traditional HFL, where model parameters are aggregated via a centralized cloud server. This shared model captures generalizable patterns across clients. The embeddings from the last hidden layer in the FL stage are extracted to serve as inputs for the later training process. After that, each client uses its local feature group to train a localized model (main net), while simultaneously feeding the embeddings from FL into a secondary network called the prune net. Finally, the overall loss is calculated as the sum of the local model's loss and a weighted prune net loss, multiplied by a learnable parameter β. This combined loss is used for backpropagation to update both local model and prune net parameters.

## A. Feature Grouping

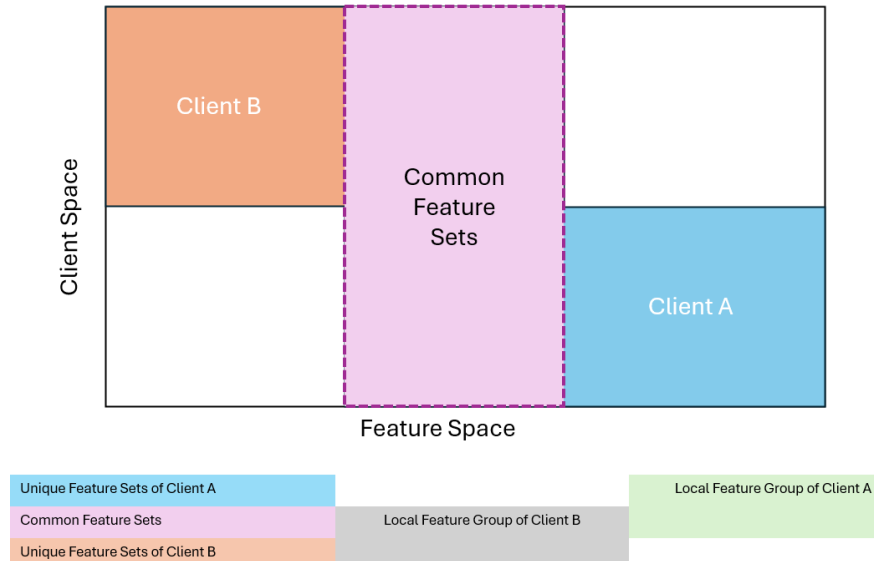

**Figure 2**: Features Grouping in two clients.

We begin by dividing the features into two distinct sets, which are then grouped into different feature groups for use in our models (**Figure 2**). We first identify the common features shared with all clients. These shared features are designated as the common feature set, as they can be used collaboratively in the FL process. The non-overlapping, client-specific features are categorized as unique features.

Ideally, the local feature group should be optimized to include only the most relevant features for each client's specific data, which may involve filtering out less informative features. However, in practice, it is often impractical to determine feature importance a priori. Therefore, in this experiment, each client's local feature group includes both the common (global) features and its unique features to ensure all available information is utilized during localized training.

## B. Federated Learning

After categorizing the features into global and local groups, FL is applied to the global feature group across all clients. Since SPC prediction is a highly imbalanced task, with the risk of developing SPC less than 10%, class weights are calculated using the method proposed in this study (Cui et al., 2019). We then extract the output from the last hidden layer of the global model to serve as input embeddings for subsequent learning stages. The motivation for using the last hidden layer outputs lies in their ability to retain rich and high-dimensional representations of the input data. Inspired by the architecture of autoencoders (Hinton & Salakhutdinov, 2006), this approach helps preserve more informative and structured features, providing a meaningful and compact representation for the subsequent learning stages.

### C. Fusing: Localized Learning and Prune Net Guidance

After retrieving the embeddings from the federated learning model, we feed them into a lightweight neural network—referred to as the prune net—comprising one layer. Since the embeddings are already trained and have converged, the prune net focuses on refining the representations further. The loss of the prune net is then multiplied by a learnable parameter $\beta$ and added to the localized learning loss, which is calculated from training a separate model using the local feature group. In this setup, the prune net loss acts as a guiding signal, dynamically influencing the optimization process by accelerating or decelerating the convergence of local model parameters.

Once the model reaches convergence, the final prediction output is derived solely from the localized learning network, which has been informed by both local features and global context.

## 3. Result

### A. Dataset

This study utilizes cancer registry data from two sources: a private Taiwanese cancer registry dataset and the publicly available U.S. SEER Program. The Taiwanese dataset comprises cancer registry records collected from five hospitals, which includes a total of 10,545 lung cancer patient records spanning the years 2011 to 2020. In contrast, the SEER dataset is based on the "Incidence - SEER Research Data, 18 Registries, Nov 2020 Sub", and includes 85,290 lung cancer patient records collected from 2000 to 2018, covering various racial and demographic groups. The study is approved by the Institutional Review Board of the Chang Gung Medical Foundation with a waiver of informed consent (201901386B0).

To facilitate model development and evaluation, each dataset is partitioned using stratified sampling based on the target into two subsets: 70% for training and 30% for testing. Moreover, to ensure robustness, each experiment is repeated across 30 different random seeds. This repeated stratified split and testing strategy allows the model to learn from a representative portion of the data while reserving an independent set for unbiased performance evaluation, which is an important consideration given the limited size of the Taiwanese dataset.

### B. Baseline Models

To highlight the importance of our proposed LF2L and prove it can effectively incorporate datasets from different sources into model training, we design a set of baseline models for comparison. The first baseline evaluates localized learning, which is trained solely on the local dataset, without incorporating any different sources data. The second baseline introduces data by implementing an HFL framework (Zhang et al., 2021), where only the common features between the two datasets are utilized. This setup reflects a straightforward but naive integration of data from the other sources without considering feature heterogeneity (Guarrasi et al., 2025). Finally, we introduce centralized learning baselines to further examine the effects of naively merging heterogeneous datasets. We pool two datasets together and retain all features by marking the unique features values as unknown. This approach aims to incorporate information from multiple sources but handles feature disparity ineffectively.

## C. Model Performance

**Table 1** presents the model performance when incorporating SEER dataset resources into the Taiwanese clinical dataset. The results demonstrate that augmenting data even from different sources can still improve prediction performance. Compared to traditional neural network training, our method achieves a higher AUROC (0.7326 ± 0.0387 vs. 0.7196 ± 0.0196, $p$ = 0.1067) and AUPRC (0.1187 ± 0.0253 vs. 0.1004 ± 0.0090, $p$ < 0.001). Even when compared to HFL, our approach significantly outperforms both AUROC (0.7326 ± 0.0387 vs. 0.7157 ± 0.0187, $p$ < 0.05) and AUPRC (0.1187 ± 0.0253 vs. 0.0953 ± 0.0072, $p$ < 0.001).

**Table 1**. Performance of Second Primary Cancer Prediction in Lung Cancer Survivors Using the Taiwanese Clinical Dataset with SEER Data as Augmentation

| Method | Model | AUROC (SD) | AUPRC (SD) |
|---|---|---|---|
| Baselines | Localized Learning | 0.7196 ± 0.0196 | 0.1004 ± 0.0090 |
| | Federated Learning | 0.7157 ± 0.0187 | 0.0953 ± 0.0072 |
| Ours | **Adding SEER dataset** | **0.7326 ± 0.0387** | **0.1187 ± 0.0253** |

On the other hand, **Table 2** presents the performance of the US-based SEER dataset when augmented with the Taiwanese local data. Compared to traditional neural network training, our method achieves a higher AUROC (0.7337 ± 0.0083 vs. 0.7219 ± 0.0113, $p$ < 0.001) and AUPRC (0.1373 ± 0.0063 vs. 0.1317 ± 0.0051, $p$ < 0.001). Compared to the federated learning baseline, our approach also outperforms in terms of AUROC (0.7337 ± 0.0083 vs. 0.7294 ± 0.0060, $p$ < 0.05) and achieves a slightly higher AUPRC (0.1373 ± 0.0063 vs. 0.1355 ± 0.0039, $p$ = 0.19), though the latter difference is not statistically significant.

Last, comparing with the performance of centralized learning by recklessly pooling two heterogeneous datasets—Taiwanese clinical data and SEER—into a single dataset. Although our method is comparable in AUPRC (0.1187 ± 0.0253 vs. 0.1171± 0.0289, $p$ = 0.82), it reaches a significant improvement in AUROC (0.7326 ± 0.0387 vs. 0.6890 ± 0.0908, $p < 0.05$) in Taiwanese dataset. Furthermore, our method results in a higher AUROC (0.6890 ± 0.0908 vs. 0.7337 ± 0.0083, $p < 0.01$) and a significantly higher AUPRC (0.1171 ± 0.0289 vs. 0.1373 ± 0.0063, $p < 0.001$) in SEER dataset.

Table 2. Performance of Second Primary Cancer Prediction in Lung Cancer Survivors Using the SEER Dataset with Taiwanese Clinical Dataset as Augmentation

| Method | Model | AUROC (SD) | AUPRC (SD) |
| --- | --- | --- | --- |
| Baselines | Localized Learning | 0.7219 ± 0.0113 | 0.1317 ± 0.0051 |
|  | Federated Learning | 0.7294 ± 0.0060 | 0.1355 ± 0.0039 |
| Ours | **Adding Taiwanese dataset** | **0.7337 ± 0.0083** | **0.1373 ± 0.0063** |

## 4. Discussion

By comparing the localized learning model with our proposed method, we observe that while localized learning leverages all available local features, it fails to utilize the large-scale dataset, which the SEER dataset contains nearly eight times more records than the Taiwanese dataset. As a result, the training data remains limited in size, constraining the model's learning capacity and leading to inferior performance. In contrast, the HFL incorporates data from two different sources, thus benefiting from increased sample size. However, due to feature alignment constraints, it only uses the common features shared between datasets, thereby discarding unique and informative clinical features from the Taiwanese dataset, such as EGFR gene mutation, and ALK gene mutation, etc. This results in a loss of critical domain-specific information and ultimately degrades predictive performance.

Our proposed method addresses both of these limitations. It preserves valuable local features from the Taiwanese dataset while leveraging the scale and diversity of the SEER dataset through the LF2L framework. Furthermore, centralized learning demonstrates how effectively our method leverages additional datasets. Centralized learning aggregates all data into a unified dataset and appears to leverage both quantity and feature diversity, but does so in a naive manner, without explicitly addressing feature-level discrepancies across datasets. The imputation process introduces sparsity and potential noise, which can impair the model's capacity to learn meaningful patterns. These observations reinforce the need for LF2L, which goes beyond simple dataset aggregation to effectively resolve feature heterogeneity in multi-source medical data.